\documentclass[a4paper,11pt,twocolumn]{article}

\usepackage{icphs2023}
\usepackage{metalogo} 
\usepackage{epstopdf}
\usepackage{multirow}

\title{What makes a good pause?\\ Investigating the turn-holding effects of fillers}
\author{Bing'er Jiang, Erik Ekstedt, Gabriel Skantze}
\organization{KTH Speech, Music and Hearing, Stockholm, Sweden}
\email{\{binger, erikekst, skantze\}@kth.se}
\begin{document}

\maketitle

\begin{abstract}
Filled pauses (or \textit{fillers}), such as \textit{uh} and \textit{um}, are frequent in spontaneous speech and can serve as a turn-holding cue for the listener, indicating that the current speaker is not done yet. In this paper, we use the recently proposed Voice Activity Projection (VAP) model, which is a deep learning model trained to predict the dynamics of conversation, to analyse the effects of filled pauses on the expected turn-hold probability. The results show that, while filled pauses do indeed have a turn-holding effect, it is perhaps not as strong as could be expected, probably due to the redundancy of other cues. We also find that the prosodic properties and position of the filler has a significant effect on the turn-hold probability. However, contrary to what has been suggested in previous work, there is no difference between \textit{uh} and \textit{um} in this regard.
\end{abstract}


\section{Introduction}

In spontaneous speech, filled pauses (or \textit{fillers}), such as \textit{uh} and \textit{um}, are frequent and are typically associated with a hesitation on part of the speaker \cite{fox_tree_interpreting_2002}. Some studies have found an association between the use of fillers and higher cognitive load in both monolog and dialog \cite{rose_um_2015}. From the perspective of coordinating turn-taking in conversation \cite{sacks100247,duncan100829,Skantze2021}, it has been suggested that fillers work as a \textit{turn-holding cue} for the listener, indicating that the current speaker is not done yet \cite{Ball1975,clark100117}. To what extent such fillers are produced intentionally for this purpose, and how much information they carry, is not yet clear \cite{corley_hesitation_2008}. 

According to \cite{clark100117}, \textit{uh} and \textit{um} are to be considered as words in the speaker's vocabulary. While these words do not have any propositional content, it is argued that fillers conform to the ``phonology, prosody, syntax, semantics, and pragmatics of English words''. In this view, they should be seen as interjections (similar to \textit{ah} and \textit{oh}), indicating that the speaker wants to keep the floor. 
In addition, \cite{clark100117} claim that \textit{uh} and \textit{um} have different meanings, where the latter signals a longer upcoming delay. This claim is based on an analysis of the pause length following these fillers in several spontaneous speech corpora. Later studies of other corpora have also found such differences \cite{rose_um_2015}, while others have not \cite{OConnell2005}. 
Another possibility is that it is not the lexical form of the filler that is important for their function in signalling delay, but rather their prosodic characteristics. If so, it would perhaps be wrong to regard them as carrying distinct symbolic meaning. However, we are not aware of any studies that have investigated this. 

A general problem with corpus-based analyses is that we can only find correlations, which do not necessarily imply causation. Thus, if we compare places with filled pauses with places without them, we do not know whether it is the filled pauses that actually cause the observed differences, or whether they were due to other factors that happened to correlate with the usage of fillers. One alternative is to perform perception experiments where stimuli are systematically manipulated, while all other factors are kept constant (e.g. \cite{Cook1970,Ball1975,Ruiter2006,Bogels2015}). However, such experiments are costly to perform on a larger scale and they are often done with a small set of made-up examples that do not necessarily reflect general distributions. In this paper, we take a third route: To train a model that makes predictions in conversation and then use that model to investigate the effects of large-scale systematic stimuli manipulation. 

Using the recently proposed \textbf{Voice Activity Projection} model \cite{vap} (\textbf{VAP}), which models the dynamics of conversation and can be used to predict the \textbf{turn-hold probability} (\textbf{THP}) (i.e., the probability that the turn will not shift to the other speaker) at any given point, we systematically remove and insert fillers in order to investigate their effects. We want to answer the following questions:

\begin{enumerate}
    \item How is the THP affected when fillers are removed from their natural occurrences?
    \item How is the THP affected when fillers are inserted at places which should be clearly turn-yielding?
    \item How do the properties of the filler (position, lexical form, pitch, intensity, duration) affect the THP?
\end{enumerate}

\section{Voice Activity Projection}
Voice Activity Projection \cite{vap}, VAP, is a predictive model of conversational dynamics developed for turn-taking research. The model objective is to continuously predict the upcoming voice activity of both speakers in a dialog. The voice activity is defined in binary terms (i.e., whether they will be speaking or not at a certain point in time), and the two speakers' future activities are jointly encoded into a discrete label that represents the next 2s of dialog. 

Similar to a language model (text-to-text), the VAP model (speech-to-activity) is optimized through MLE (cross-entropy). The model processes raw spoken dialog audio and outputs a likelihood distribution over each of the discrete activity labels. This likelihood can then be used for controlling the turn-taking in a spoken dialog system, or as a tool to analyze the likelihood of turn-taking events in recorded dialogs. In previous work, it has been shown that the model is sensitive to prosodic cues \cite{ekstedt_how_2022}. In this work, we utilize the VAP model to investigate the effect of a filler has on how long the current turn holds until the other speaker takes the turn.

The activity projection is defined by eight time window bins of a 2s window,
four for each speaker (0.2s, 0.4s, 0.6s, 0.8s respectively), which indicates if the speaker is active or not in that time window. The various combinations of these bins result in $2^8$ = 256 discrete activity labels. 
While these labels represent various turn-taking-related events (such as turn-shifts and backchannels), we are here primarily interested in calculating the probability of who will be the next speaker. To do this, we take a weighted average over the probability assigned to each label, focus on the bins that cover our region of interest (0-600ms), and compare the contributions for each speaker. The result is a value between 0 and 1 that represents the probability that the current speaker will continue, defined as \textbf{turn-hold probability (THP)}. 
In our study, we apply the model on dialogs and get incremental predictions of THP.

The model used in this experiment is a stereo version of the original VAP model that operates on two separate waveforms (one for each speaker) and is trained on subsets of the Fisher part 1 and the Switchboard corpus \cite{fisher,swb}. We define a held-out test set containing 577 sessions from the dialog act annotated subset \cite{swda}, swda, of the switchboard corpus to be used in the subsequent analysis.

\section{Experiment 1: excluding fillers}

\begin{figure}[t]
\begin{center}
\includegraphics[width=0.47\textwidth]{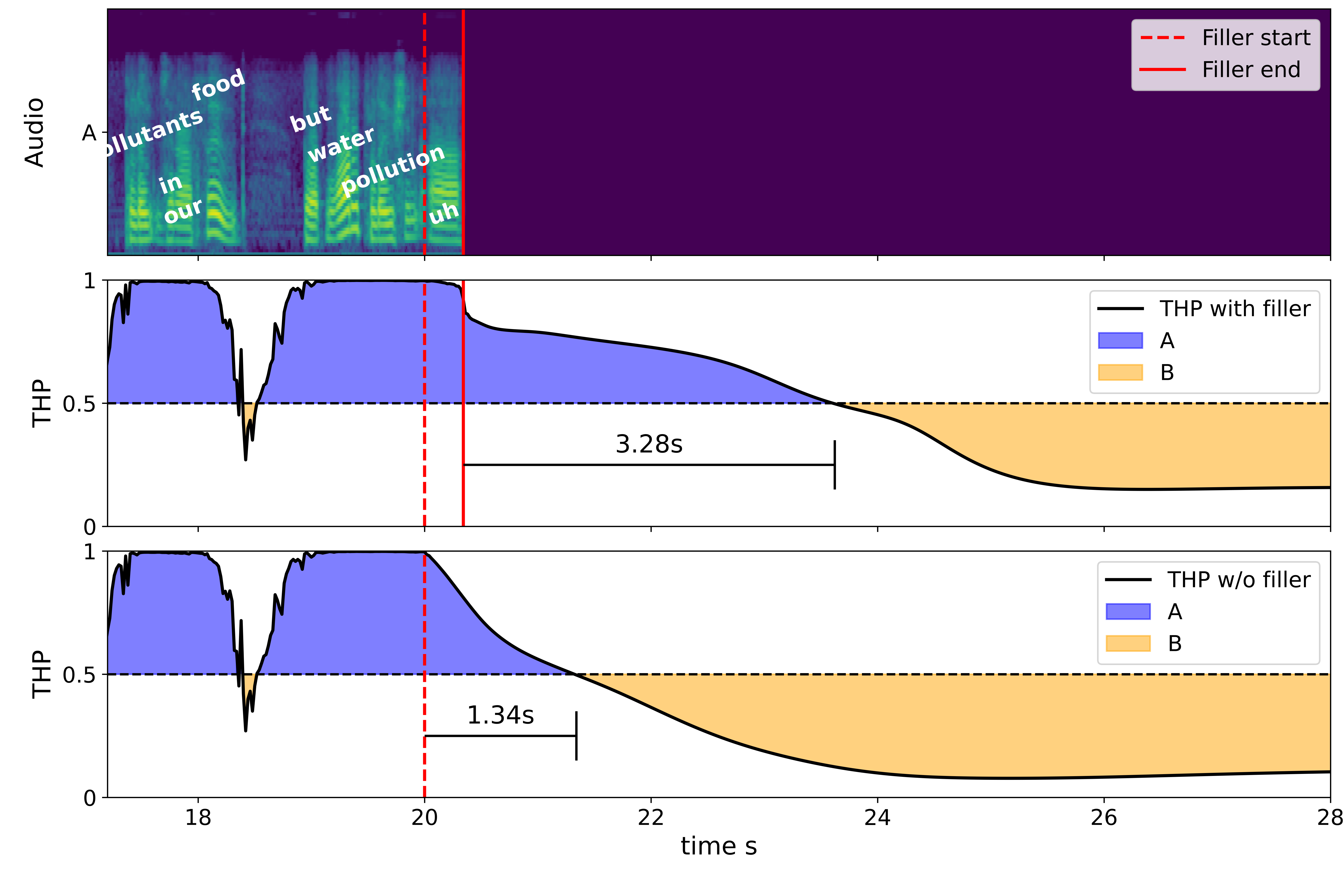}
\caption{The THP following an utterance ending with \textit{uh}. Top: the spectrogram of the spoken utterance; Middle: the THP with the filler; Bottom: the THP without the filler. 
The blue/yellow areas corresponds to A/B being the most likely next speaker. The black number bars represent the duration of silence before a turn-shift prediction (THP falls under 0.5). }\label{fig:example-omission}
\end{center}
\end{figure}

We evaluate the effect of fillers through examining how `soon' the VAP model predicts a turn shift after the end of a speech segment.
In Experiment 1, we investigate the effect of excluding a valid filler present in the data. 
As illustrated in Figure \ref{fig:example-omission}, we apply the model to 20s of dialog (a pair of data samples including and excluding the filler), followed by 10s of artificial silence, and track the THP over the silence segment. While a pause of 10s is very unlikely in real life conversations, we use a longer length anyways to fully examine the prediction of the turn shift. The time \textit{t} where the THP drops below 50\% is considered the position of turn shift.

We select filler candidates \textit{uh} and \textit{um}, based on word-level transcriptions, and define three criteria in order to consider them valid fillers: 1) the duration of the candidates must be longer than 0.2s; 2) the fillers should be followed by a pause of at least 0.2s of the same speaker and `isolated' from activity of the listener of at least 1s, before and after the filler; 3) there should be at least 20s of past dialog context (for the model to make its predictions). This results in 5316 valid fillers from the 577 test sessions.


We use \textit{Survival Analysis} to examine the results, as some cases may not reach the turn shift predicted by the model even at the end of the 10s silence. The \textit{Survival Analysis} includes a \textit{censoring} status to identify whether the expected event has happened or not (\textit{uncensored} vs. \textit{censored}). In our case, dialogs that are not predicted to reach turn shift by the end of the silence are marked as \textit{censored}.

\begin{figure}[t]
\begin{center}
\includegraphics[width=0.45\textwidth]{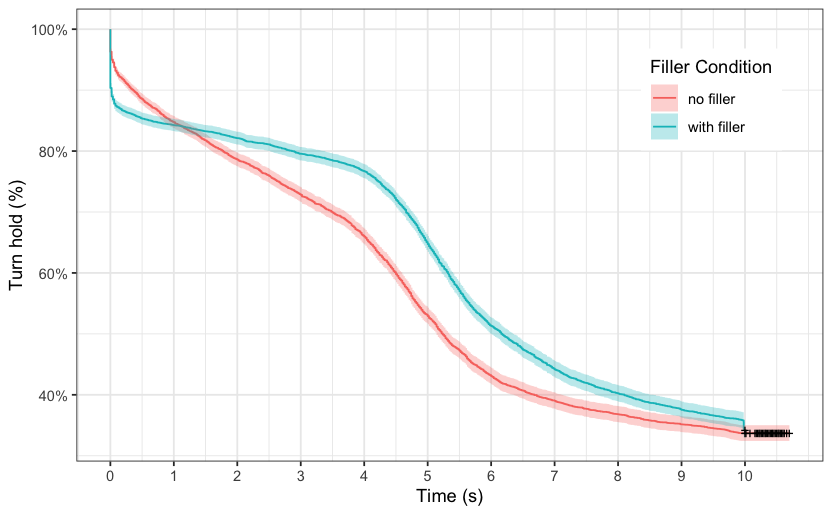}
\caption{Kaplan-Meier curve of turn hold predictions through 10s of silence for Experiment 1 (excluding fillers), with or without fillers. }
\label{fig:km-exp1}
\end{center}
\end{figure}

In order to examine whether the filler makes a difference, we plot the Kaplan-Meier curve of the two conditions. Figure \ref{fig:km-exp1} shows the proportion of data samples that keep the hold (y-axis) as a function of time (x-axis). Note that the y-axis does \textit{not} reflect the THP; instead, it refers to percentages of cases where the THP are above 50\%, i.e., percentages of data that are likely to keep the hold at time \textit{t}. 

Figure \ref{fig:km-exp1} shows that the filler indeed makes a difference in holding the turn, confirmed by the log-rank test (\textit{p} < 0.001). Moreover, the effect changes throughout time. In particular, the filler does not help holding the turn at the very beginning of the silence, as shown by the drastic drop of the green line. At the later stage of the silence (after 1.2s), the effect of keeping the hold by the filler becomes more prominent, reaching the greatest between 4s and 5s, and then decreases until the end of the silence.  

While there is a clear effect of the filler, it is interesting to note that the effect is perhaps not as big as expected; the turn hold effect is quite strong even with the filler removed. This is likely due to other redundant turn-holding cues present before the filler (prosody, syntax, etc). 

\section{Experiment 2: inserting fillers}
While Experiment 1 indeed shows a turn-holding effect of existing fillers, it is unclear how strong the effect is, given the confounded turn-holding context. 
In Experiment 2, we further investigate the effects of fillers in contexts that are clearly turn-yielding. Specifically, 
we insert fillers at the end of yes/no questions and examine the difference in THP between the original data samples and the ones with inserted filler at the end.

We define valid utterances by the following conditions: we first select all utterances that are labeled as a yes/no-question dialog act according to the \textit{swda} annotation \cite{swda}. The original utterance annotations may include multiple dialog act segments in a single utterance and we omit those that do not end with a yes/no-question dialog act. Valid utterances must be followed by a pause longer than 0.5s (from the same speaker) and `isolated' from activity of the listener of at least 0.5s, before and after the end of the utterance. 
The candidates with a context duration of less than 20s are omitted. The final condition is that the model assigns a turn-shift in the first 5s of silence when 20s of context up and including the utterances are processed. This approach result in 245 valid turn-shift utterances.

\begin{figure}[t]
\begin{center}
\includegraphics[width=0.47\textwidth]{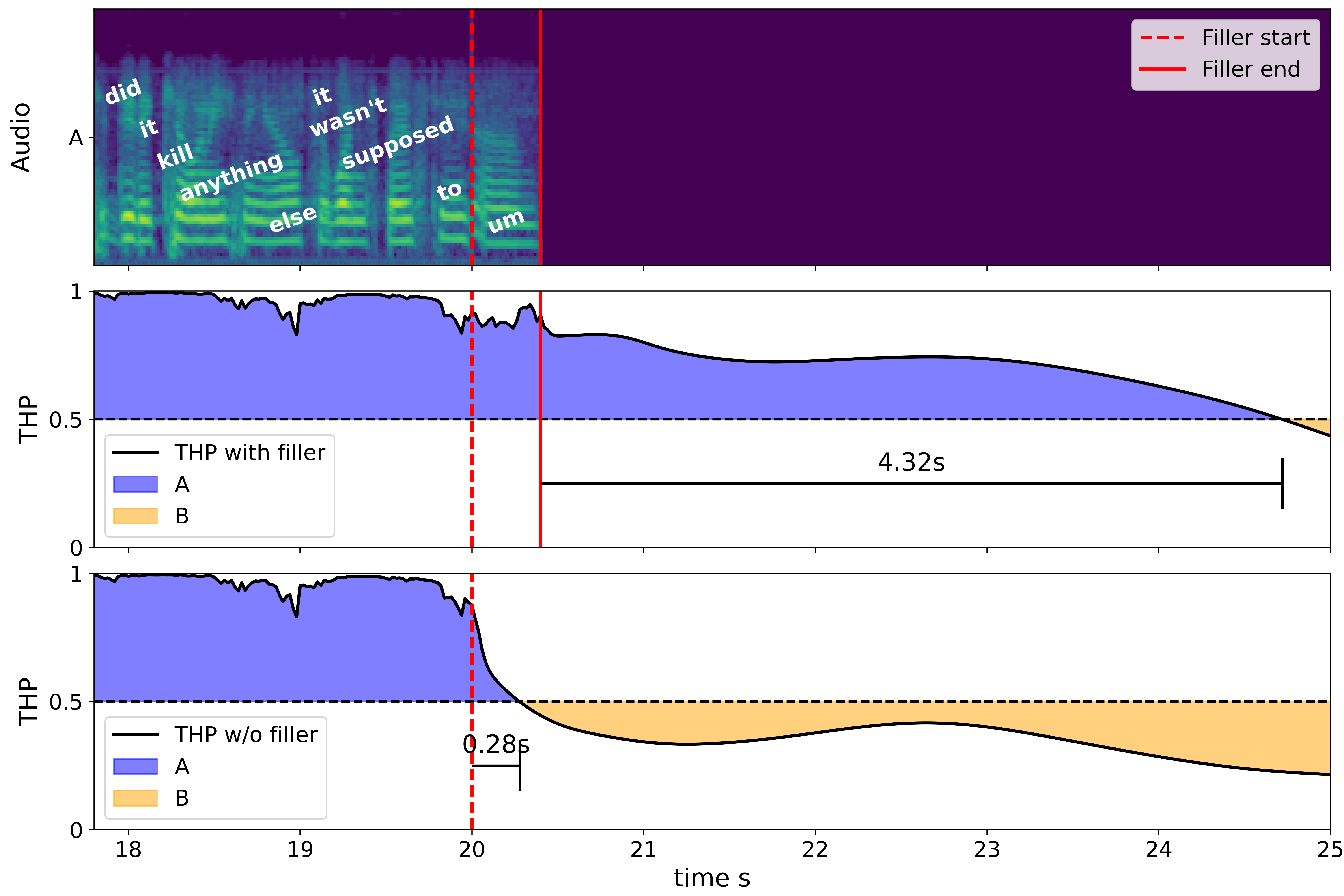}
\caption{The THP following a yes/no question. Top: the spectrogram of the spoken utterance; Middle: the THP with the insertion of the filler \textit{um}; Bottom: the THP of the original utterance.}
\label{fig:example-addition}
\end{center}
\end{figure}

For each valid utterance we select all the valid fillers, from the same speaker and session, as candidates to add to the utterance (1436 total filler candidates). Similarly to Experiment 1, we create paired data samples constituting 20s of dialog context, up to and including the utterances (with or without the added filler), followed by 10s of silence. This is illustrated in Figure~\ref{fig:example-addition}. 

\begin{figure}[t]
\begin{center}
\includegraphics[width=0.45\textwidth]{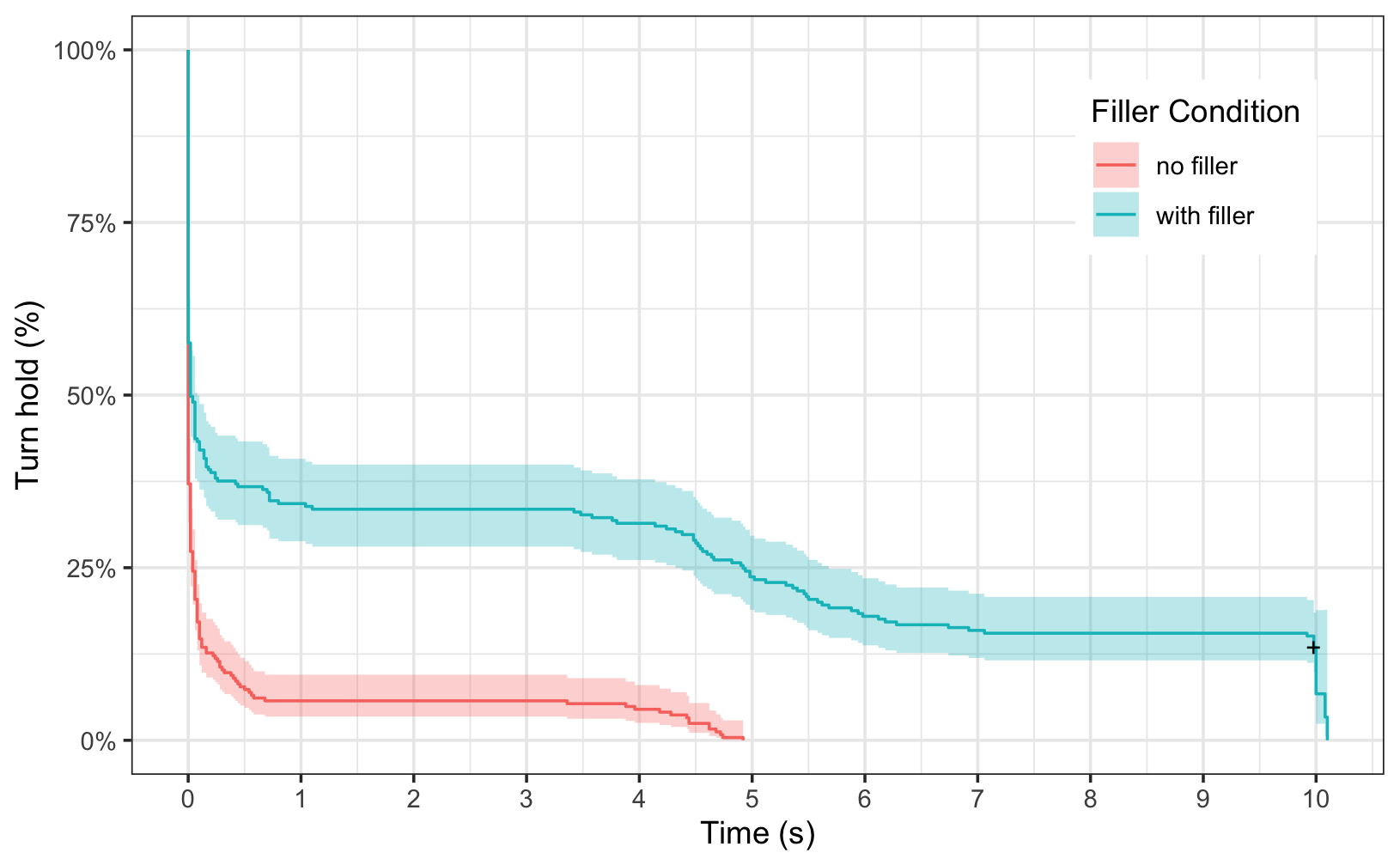}
\caption{Kaplan-Meier curve of turn hold predictions through 10s of silence for Experiment 2 (inserting fillers after yes/no questions), with or without fillers.}
\label{fig:exp2-km}
\end{center}
\end{figure}


Following the same preprocessing as described in Experiment 1, we plot the Kaplan-Meier curve, as shown in Figure \ref{fig:exp2-km}. The figure shows that inserting a filler consistently lengthens the hold, and the two curves differ significantly from the log-rank test (\textit{p} < 0.001). As can be seen, for the original yes/no questions, the model clearly predicts a fairly quick turn shift in most cases. While adding a filler does not flip all of them to turn-holds, the fillers seem to have a stronger effect here than in Experiment 1, and so the model clearly is sensitive to them. 

\section{Experiment 3: prosodic and lexical effects}

Experiment 3 examines how prosodic and lexical properties of the filler affect the turn hold. 
For capturing prosodic information, we include by-speaker standardized mean \textbf{pitch}, by-speaker standardized mean \textbf{intensity} and filler \textbf{duration} on the log scale. We also examine whether the \textbf{lexical form} (\textit{uh} or \textit{um}), as well as the \textbf{position} of the filler in the current dialogue act (\textit{mid} or \textit{start}) has any influence on the THP. All continuous variables (prosody) are standardized by subtracting the mean and divided by two standard deviations. Note that the pitch and intensity variables are normalized with regard to each speaker, while duration is not.



We fit a Cox proportional hazards model on data with fillers from Experiment 1 to examine the joint effects from the five factors. 
The independent variables are the five factors and an interaction term between F0 and the lexical form, and the dependent variables are time reaching the turn shift derived from the VAP model and the censor status. 

Table \ref{tab:cox_results}, summarizing the Cox model, shows that except for the filler type, all four other variables are statistically significant. The coefficients reflect the change of hazards function (here, the probability of the turn shift): a negative coefficient estimate corresponds to less `hazard' of turn shift (higher probability of turn hold), while a positive coefficient estimate indicates an increased probability of turn shift. The exponentiated coefficients are termed as \textit{hazards ratio}, which represents the difference in the hazard of an event at any given time.

The results from Table \ref{tab:cox_results} reveals that the following factors help keep the hold (decrease the `hazard'): higher pitch, start of the dialog act, higher intensity and longer duration. Moreover, while the lexical form of the filler itself does not significantly affect the hold, it has an indirect influence through pitch: \textit{uh} has higher pitch and thus results in longer holds.

Specifically, if a speaker raises their voice by 0.5 standard deviation (SD) of their pitch range, the probability of holding the turn determined by the model will increase by 51.6\% (here, the change of hazards of turn shift from \textit{hazards ratio} is 0.484 - 1 = -0.516). Increasing 0.5 SD of the intensity, however, only raises the probability by 12.1\%.
The duration also affects the turn hold: if the filler is lengthened by 0.5 SD, the probability is raised by 11.2\% .
In terms of the position, if a filler is at the beginning of the dialog act, it tends to raise the probability by 26.4\% compared to one in the middle.

\begin{table}[t]
\centering
\begin{tabular}{ lrrrr } 

\hline
                        &coef& coef(exp)& SE  &    Pr(>|z|)  \\  
\hline
F0           &     -0.725 &  0.484 & 0.246 & \textbf{0.003} \\
Intensity     &    -0.127  & 0.879 & 0.035 &  \textbf{0.0003} \\
Lex\textsubscript{\textit{um}}          &        -0.077  & 0.925 & 0.050 & 0.12   \\ 
Duration      &     -0.118 &  0.888 & 0.037 &  \textbf{0.001}\\
Pos\textsubscript{\textit{mid}}          &   -0.305 &  0.736 & 0.065 & \textbf{<0.0001} \\
F0:Lex\textsubscript{\textit{um}} & -1.237 &  0.237 & 0.290 &  \textbf{0.007} \\ 
\hline

\end{tabular}
\caption{Model summary of the Cox regression model. Bold \textit{p} values are significant. }
\label{tab:cox_results}
\vspace{1em}
\end{table}

\section{Discussion and Conclusions}

In this study, we use computational modelling 
to evaluate the effects of fillers on holding a turn, and examine a variety of prosodic, positional and lexical properties of the fillers. 
Experiment 1 shows that excluding existing fillers indeed results in shorter holds, while the effect is perhaps not as strong as expected. This is likely due to the redundancy of other turn-holding cues. 
Experiment 2 further reveals that fillers have a strong turn-holding effect when added to clearly turn-yielding contexts (yes/no questions). 
Experiment 3 shows that a number of properties of the filler contribute to the turn-hold effect: higher pitch, higher intensity, and longer duration of the filler are associated with longer turn hold. Moreover, utterance-initial fillers have a stronger turn-hold effect than mid-utterance fillers. However, there is no significant difference between the lexical forms \textit{uh} and \textit{um}. Nevertheless, the two types of fillers still indirectly affects the THP through increasing the pitch of the filler. Thus, instead of the lexical forms, it is their phonetic realization that makes a difference, as \textit{um} includes a nasal which associates with lower pitch compared to the oral \textit{uh}. This speaks against the interpretation given by \cite{clark100117}, where \textit{uh} and \textit{um} are seen as different words with different meanings. 

\section{Acknowledgements}
This work was supported by Riksbankens Jubileumsfond (RJ), through the project \textit{Understanding predictive models of turn-taking in spoken interaction} (P20-0484), as well as the Swedish Research Council, through the project \textit{Prediction and
Coordination for Conversational AI} (2020-03812).

\bibliographystyle{IEEEtran}
\bibliography{icphs2023}


\end{document}